\documentclass[11pt]{article}
\usepackage{coling2016}
\usepackage{times}
\usepackage{url}
\usepackage{latexsym}
\usepackage{graphicx}
\usepackage{subfigure}
\usepackage{array}
\usepackage{amsmath}
\usepackage{amssymb}
\usepackage{mathrsfs}
\usepackage{color}

\title{A Novel Bilingual Word Embedding Method for Lexical Translation \\ Using Bilingual Sense Clique }

\author{ Rui Wang$^{1}$, Hai Zhao$^{1}$, Sabine Ploux$^{2}$, Bao-Liang Lu$^{1}$, Masao Utiyama$^3$ and Eiichiro Sumita$^3$\\
	$^1$ Shanghai Jiao Tong University, Shanghai, China\\
	$^2$Centre National de la Recherche Scientifique, CNRS-L2C2, France\\
	$^3$National Institute of Information and Communications Technology, Kyoto, Japan\\
	wangrui.nlp@gmail.com, \{zhaohai, blu\}@cs.sjtu.edu.cn, \\
	sploux@isc.cnrs.fr, \{mutiyama, eiichiro.sumita\}@nict.go.jp \\
}

\date{}

\begin{document}
\maketitle
\begin{abstract}
 Most of the existing methods for bilingual word embedding only consider shallow context or simple co-occurrence information. In this paper, we propose a latent bilingual sense unit (Bilingual Sense Clique, BSC), which is derived from a maximum complete sub-graph of pointwise mutual information based graph over bilingual corpus. In this way, we treat source and target words equally and a separated bilingual projection processing that have to be used in most existing works is not necessary any more. Several dimension reduction methods are evaluated to summarize the BSC-word relationship. The proposed method is evaluated on bilingual lexicon translation tasks and empirical results show that bilingual sense embedding methods outperform existing  bilingual word embedding methods.
\end{abstract}

\section{Introduction}
\label{sec:intro}
Bilingual word embedding can enhance many cross-lingual natural language processing tasks, such as word translation, cross-lingual document classification and Statistical Machine Translation (SMT) \cite{mikolov2013exploiting,lauly2014autoencoder,kovcisky-hermann-blunsom:2014:P14-2,gouws2015bilbowa,gouws-sogaard:2015:NAACL-HLT,vulic2015bilingual,mogadala-rettinger:2016:N16-1,zhang2016building}. It can be considered as a \emph{cross-lingual} projection \cite{upadhyay2016cross} of monolingual word embedding \cite{mikolov2013distributed,pennington-socher-manning:2014:EMNLP2014}. According to the \emph{cross-lingual} projection step, there are mainly three types of bilingual embedding methods. 

1) Each language is embedded separately at first, and transformation of projecting one embedding onto the other are learned using word translation pairs then. Mikolov et al. \shortcite{mikolov2013exploiting} propose a linear projection method, which is further extended by \cite{xing-EtAl:2015:NAACL-HLT} with a normalized objective method and \cite{faruqui-dyer:2014:EACL,lu-EtAl:2015:NAACL-HLT} with canonical correlation analysis. 

2) Parallel sentence/document-aligned corpora are used for learning word or phrase representation directly \cite{ettinger-resnik-carpuat:2016:N16-1}. Recently, Neural Network (NN) based projection methods are widely used for this kind of embedding \cite{hermann2013multilingual,hermann-blunsom:2014:P14-1,lauly2014autoencoder,zhang-EtAl:2014:P14-11,gao-EtAl:2014:P14-1}. 

3) Monolingual and bilingual objectives are optimized jointly  \cite{klementiev2012inducing,zou-EtAl:2013:EMNLP,luong2015bilingual,shi-EtAl:2015:ACL-IJCNLP1,Vulic-acl16}. Typically, a small parallel sentence-aligned corpus and a large monolingual corpus are needed \cite{gouws2015bilbowa,coulmance-EtAl:2015:EMNLP}.

Meanwhile, most of these methods use bag-of-word,  $n$-grams, skip-grams or other local co-occurrence to exploit word relationship, and then use various cross-lingual projection methods to summarizing the relationship. One question arises: can we construct the cross-lingual relationship before the projection step?

As we know, sense gives more exact meaning formulization than word itself and graph can gain more global relationship than contextual relationship. Motivated by these,  we propose a Bilingual Sense Clique (BSC), which is extracted from bilingual Point-wise Mutual Information (PMI) graph using parallel corpora. BSC can be viewed as a automatic gained bilingual synset (a small group of synonyms labeled as concept) in WordNet  \cite{miller1990introduction} and plays a role of minimal unit for bilingual sense representation. Several dimension reduction methods are used for summarizing BSC-word matrix into lower dimensions vectors for word representation. This work extends previous monolingual method \cite{Ploux:2003:MMS:873743.873744,Isc03lexicalknowledge}, which inconveniently need bilingual lexicons or synonyms for bilingual mapping.  In comparison with previous bilingual graph-based semantic model which focus on SMT \cite{bgsm}, we apply several dimension reduction methods and focus on lexical translation.

The remainder of this paper is organized as follows. Section \ref{sec:cli} will introduce the bilingual sense clique. Dimension reduction methods are applied to clique-word relationship summarizing in Section \ref{sec:sp}. The proposed method is evaluated in lexical translation task in Section \ref{sec:lexical}. We will discuss and analyze the related work in Section \ref{sec:related}. The last section will conclude this paper.

\section{Bilingual Sense Clique}
\label{sec:cli}

\subsection{Bilingual PMI Graph Constructing}
\label{sec:initial}
To construct a bilingual graph from a corpus, words are formally considered as nodes and co-occurrence relationships of words are considered as the edges of graph. That is, if two words (either source or target word) $n_i$ and $n_j$, if they are in the same bilingual sentence (a combination of source sentence and its aligned target sentence), they are called co-occurrences for each other and an edge are connected between them. The Edge Weight ($EW$) is defined as a PMI based co-occurrence relationship,

\begin{equation}
		\label{eq:alpha3}
		EW = \frac{{Co(n_i, n_j)}} {fr(n_i)\times fr(n_j)} ,
\end{equation}%
where $Co(n_i, n_j)$ is the co-occurrence counting of $n_i$ and $n_j$ and $fr(n)$ stands for how many times $n$ occurs in corpus. We discard the edges with low $EW$ to filter\footnote{In IWSLT task, the threshold is set as $3 \times 10^{-4}$.} the edges which are unnecessarily connected with stop words such as \emph{of, a, the} in English.

\subsection{Clique Extraction}
\label{sec:ec}
In graph theory, clique is a subset of nodes of an undirected graph such that its induced subgraph is complete \cite{clique}. For the rest of papers, all cliques are referred to the maximal clique. That is, every two distinct nodes in the clique are adjacent and the clique does not exist exclusively within a larger clique.

\begin{figure}[htbp]
	\centering
	\includegraphics[width=0.65\textwidth]{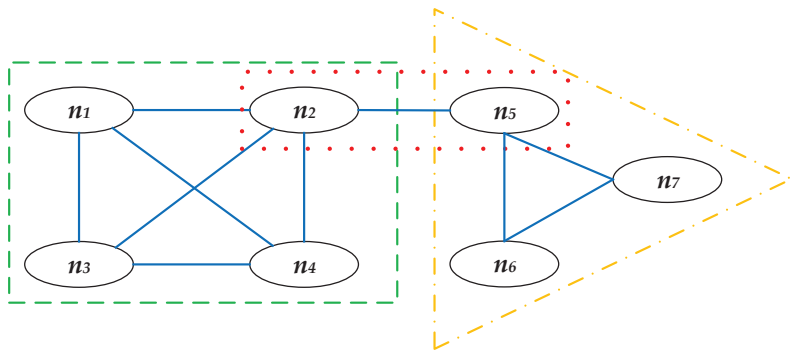}
	\caption{Three cliques are formed with $\{n_1, n_2, n_3, n_4\}$ (in green), $\{n_2, n_5\}$ (in red) and $\{n_5, n_6, n_7\}$ (in yellow).}
	\label{tab:fig3}
\end{figure}

Figure \ref{tab:fig3} illustrates an example on how to define cliques in an undirected graph. Figure \ref{tab:fig3} shows that \{$n_1, n_2, n_3, n_4$\}, \{$n_2, n_5$\} and \{$n_5, n_7, n_8$\} form three cliques, respectively. However, \{$n_1,n_3,n_4$\} is not a clique, as it is a subset of \{$n_1, n_2, n_3, n_4$\}.

Clique extraction is a non-trivial task. Some nodes in the graph need to be pruned before clique extraction due to two reasons: \uppercase\expandafter{\romannumeral1}) Sparsity of the graph. There are some nodes that do not connect any words to be embedded, so these nodes actually have no direct impact over clique extraction or further word representations. \uppercase\expandafter{\romannumeral2}) Computational complexity. Graph problems are usually associated with high computational complexity, such as finding all cliques from a graph (\emph{Clique Problem}).  The \emph{Clique Problem} in general has been shown \emph{NP-complete} \cite{karp1972reducibility}. Without any pruning, it is time consuming or even impossible to find the cliques from the whole graph built even from not a so large corpus (such as WMT11 and IWSLT-2014 which contains around 100K sentences).

For a word $n$, only its co-occurrence nodes $n_{co}$ are defined as indeed useful. The set of nodes $\{n_{co}\}$ with their weighted edges form an extracted graph for further cliques extraction for $n$. The number of nodes $\left| N_{extracted} \right|$, in the extracted graph $G_{extracted}$, is computed by

\begin{equation}
	\label{eq:nodes}
	\left| N_{extracted} \right| =  \left| \bigcup_{\forall i,j} \{n_{ij}\} \right|.
\end{equation}

Taking the example in Figure \ref{tab:fig3} again, only the extracted graph containing $\{n_1, n_2, n_3, n_4, n_5\}$ is used for clique extraction, if we want to embed $n_2$. Other nodes, such $\{n_6, n_7\}$, are not considered because they do not have connecting with $n_2$. This makes the clique extraction efficient.  In practice, $\left| N_{extracted} \right|$ is much smaller than $\left| V \right|$ (vocabulary size of bilingual corpus). For a typical corpus (IWSLT in Section \ref{sec:lexical}), $\left| N_{extracted} \right|$ is around 371.2 on average and $\left| V \right|$ is 162.3K. Thus the clique extraction in practice is quite efficient as it works over a quite small sized graph.

Clique extraction may follow a standard routine in \cite{clique}. \textbf{As a clique in this paper is to represent a bilingual sense unit of a word, it is called Bilingual Sense Clique (BSC)}. For  IWSLT-2014 French-English parallel corpus, taking the English word \emph{language} as an example, three its typical BSCs are shown in Table \ref{tab:bcc}.

\begin{table}[h]
	\begin{center}
		\begin{tabular}{l|l}
				\hline
				\hline
				BSC-1 & {\color{red}\textit{barri{\`e}res}} (barriers), {\color{blue}\textit{cultural}}, {\color{red}\textit{culturelles}} (cultural), {\color{blue}\textit{language}}, {\color{blue}\textit{mobility}}\\
				
				\hline
			BSC-2 	& {\color{blue}\textit{english}}, {\color{blue}\textit{language}}, {\color{red}\textit{langue}} (language), {\color{red}\textit{parle}} (speaks), {\color{blue}\textit{speak}}\\
				
				\hline
			BSC-3	& {\color{blue}\textit{express}}, {\color{red}\textit{exprimer}} (express), {\color{red}\textit{langage}} (language), {\color{blue}\textit{language}} \\
				\hline
				... & ... \\
				\hline
				\hline
			\end{tabular}
	\end{center}
	\caption{\label{tab:bcc} English words are shown as blue and French as red. Words in parentheses are corresponding English translations.}
\end{table}

In Table \ref{tab:bcc}, the first BSC containing \textit{cultural}, \textit{barri{\`e}res} et al. may indicate the sense of `language culture'. The second one containing \textit{speak}, \textit{english} et al. may indicate the sense of `spoken languages'. Though noise or improper connections also exist at the same time, different senses will naturally result in roughly different BSCs from our empirical observations. BSCs can be regarded as a loose but automated grained synset in WordNet  \cite{miller1990introduction}, as only strongly related words can be nodes in the same clique that possess full connections with each other. Therefore \emph{BSC-Word} relationship can obtain more exact semantic relation between word and its senses, in comparison with only using shallow bag-of-word or context sliding window relationship.

\section{Dimension Reduction}
\label{sec:sp}
To obtain concise semantic vector representation, three dimension reduction methods are introduced. Both Principal Component Analysis (PCA) \cite{person1901lines} and Correspondence Analysis (CA) \cite{hirschfeld1935connection} can summarize a set of possibly correlated variables into a smaller number of variables which is also called principal components in PCA. All these variables are usually in a vector presentation, therefore the processing is performed as a series of matrix transformations. The importance of every output components may be measured by a predefined variable, which is variance in PCA and is called inertia in CA. The most difference between them is that CA treats rows and columns equivalently. For either method, we can select top ranked components according to their importance measure so that dimension reduction can be achieved.

\subsection{Principal Component Analysis}
In this paper, PCA is conducted over the clique-word matrix constructed from the relation between BSCs and words. An initial correspondence matrix $X=\{x_{ij}\}$ is built, where $x_{ij}$ = 1 if the BSC in row $i$ contains the word in column $j$, and 0 if not. Take the example in Table \ref{tab:bcc} again, part of BSC-word initial matrix is shown in Table \ref{tab:matrix}.

\begin{table}[htbp]
	
	\begin{center}
		\begin{tabular}{l|r | r | r | r}
			\hline
			\hline
			BSC-Word &  \textit{language} & \textit{langue} & \textit{cultural} & ...\\
			\hline
			BSC-1  & 1 & 0 & 1 \\
			BSC-2 & 1 & 1 & 0\\
			BSC-3 & 1 & 0 &  0\\
			... &  &  &  \\
			\hline
			\hline
		\end{tabular}
		\caption{\label{tab:matrix}Part of BSC-word initial matrix for the example in Table \ref{tab:bcc}.}
	\end{center}
	
\end{table}

We wish to linearly transform this matrix  $X$ (whose vectors are normalized to zero mean), into another matrix $Y = PX$, whose covariance matrix $C_Y$ maximises the diagonal entries and minimises the off-diagonal entries (diagonal matrix).

	\begin{equation}
			\label{eq:cy}
			C_Y = \frac{(PX)(PX)^T}{n-1} = \frac{P(XX^T)P^T}{n-1} = \frac{PSP^T}{n-1},
	\end{equation}%
		where $S = XX^T = EDE^T$. $E$ is an orthonormal matrix whose columns are the orthonormal eigenvectors of $S$, and $D$ is a diagonal matrix which has the eigenvalues of $S$ as its (diagonal) entries. By choosing
		the rows of $P$ to be the eigenvectors of $S$, we ensure that $P = E^T$ and vice-versa. The principal components (the rows of $P$) are the eigenvectors of $S$, and in order of `importance'.

\subsection{Correspondence Analysis}	
Similar PCA, CA also determines the first $n$ factors of a system of orthogonal axes that capture the greatest amount of variance in the matrix. In this paper, PCA and CA use the same initial BSC-word matrix as original matrix $X$. Normalized correspondence matrix $\mathcal{P}=\{p_{ij}\}$ is directly derived from $X$, where $p_{ij}$ = $x_{ij}/N_X$, and $N_X$ is grand total of all the elements in $X$. Let the row and column marginal totals of $\mathcal{P}$ be $r$ and $c$ which are the vectors of row and column masses, respectively, and $D_r$ and $D_c$ be the diagonal matrices of row and column masses. Coordinates of the row and column profiles with respect to principal axes are computed by using the Singular Value Decomposition (SVD).

Principal coordinates of rows $\mathcal{F}$ and columns $\mathcal{G}$:

\begin{equation}
		\label{eq:chi}
		\mathcal{F} = {D_r}^{-\frac{1}{2}}U \Sigma, \ \ \mathcal{G} = {D_c}^{-\frac{1}{2}}V \Sigma,
\end{equation}%
where $U$, $V$ and $\Sigma$ (diagonal matrix of singular values in descending order) are from the matrix of standardized residuals $S$ and the SVD,

\begin{equation}
\label{eq:chi}
\begin{array}{ll}
S = U \Sigma V^{*} = {D_r}^{-\frac{1}{2}}(P-rc^{*}){D_c}^{-\frac{1}{2}},
\end{array}
\end{equation}%
where $*$ denotes conjugate transpose and $\ U^{*}U = V^{*}V = I$.

By above processes, CA projects BCCs ($\mathcal{F}$) and words ($\mathcal{G}$) onto semantic geometric coordinates as vectors. Inertia $\chi^2 / N_X$ is used to measure semantic variations of principal axes for $\mathcal{F}$ and $\mathcal{G}$.

\subsection{Neural Network}
\label{sec:nn}			
We apply the basic Continuous Bag-of-Words (CBOW) Model structure in word2vec \cite{mikolov2013efficient}, and BSC is considered as bilingual Bag-of-Words. As mentioned in Section \ref{sec:ec}, because BSCs are from an extracted graph $G_{extracted}$, we do not have sufficient BSCs for effective NN training. Namely, due to the data sparse  (there is only some hundreds of BSCs for each word on average), NN based method cannot not be directly applied to this sized BSC-word matrix summarization as PCA/CA does. We thus try to directly extract all of the BSC from the whole graph $G$. That is, all the  BSCs of each word (we discard words whose frequency is less than 5) in the corpus are pre-computed, and they are then as a whole used as the input of CBOW Model. The window size of CBOW is set as 8, which is default setting of word2vec. We discard the BSCs containing more than 8 words and set the projections of the missing words to zero for BSCs containing less than 8 words.

\subsection{Visualization}
Take \textit{language} as example again, we show its top-two-dimension representation by CA and NN in Figure 2. PCA is applied to summarize NN embeddings on  two-dimension (only for this visualization). Only the nearest words are shown. As shown in Figure 2, the proper French translations \textit{langue} and \textit{langage} are close to \textit{language}. The other words surrounding \textit{language} are quite different.

\begin{figure}[htbp]
	\label{fig:tu4}
	\centering
		\subfigure[CA]{
			\begin{minipage}[t]{0.5\linewidth}
				\includegraphics[width=1\textwidth]{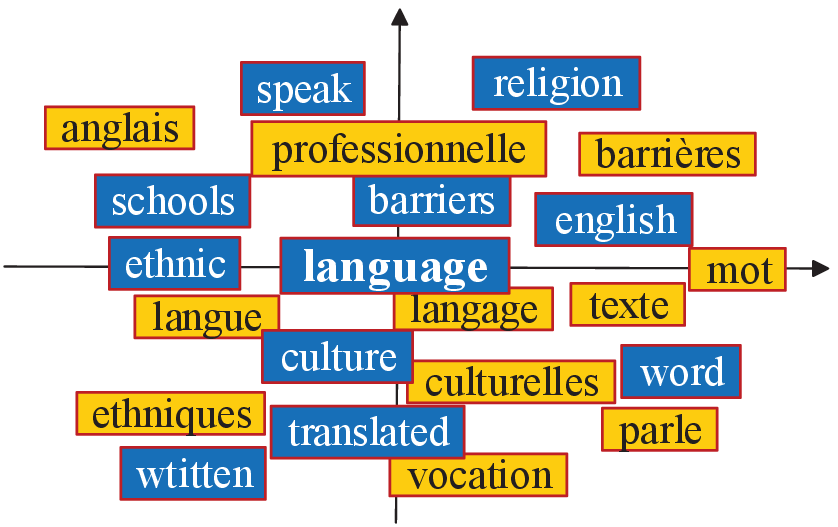}
			\end{minipage}}
			
		\subfigure[NN]{
			\begin{minipage}[t]{0.5\linewidth}
				\includegraphics[width=1\textwidth]{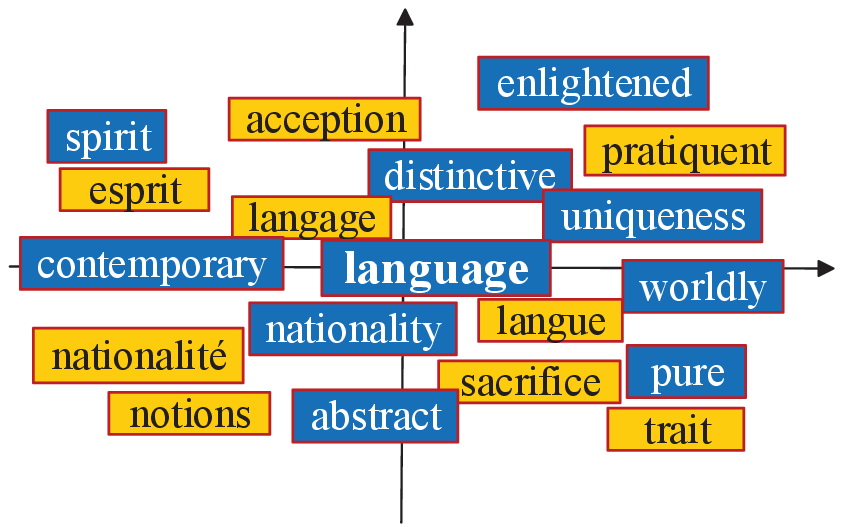}
			\end{minipage}}
				
	\caption{Visualization of word representation.}
\end{figure}

\section{Lexical Translation}
\label{sec:lexical}

\subsection{Task Description}				
Following the previous lexicon translation settings \cite{mikolov2013exploiting}, 6K most frequent words from the WMT11 Spanish-English (Sp-En) data\footnote{\url{http://www.statmt.org/wmt11/}} are translated into target languages by the online Google Translation (individually for En and Sp). Since Mikolov method requires translation-pairs for training, they used the first 5K most frequent words to learn the ``\emph{translation matrix}", and the remaining 1K words were used as a test set. The proposed method only use parallel sentences for training, so we use the first 5K most frequent words for dimension tuning and the 1K test-pairs for evaluation. To translate a source word, we finds its $k$ nearest target words with Euclidean distance, and then evaluate the translation precision $P@k$ as the fraction of target translations that are within the top-$k$ returned words. We also evaluate these methods on IWSLT-2014 French-Engliash (Fr-En) task\footnote{\url{https://wit3.fbk.eu}}, with the same setting as WMT11 task.  The corpus statistics are shown in Table \ref{tab:dataset}.

\begin{table}[htbp]
	
	\begin{center}
		\begin{tabular}{l|r | r }
			\hline
			\hline
			Corpus &  WMT & IWSLT \\
			\hline
			training  & 132K & 178K \\
			dev & 5K & 5K \\
			test & 1K & 1K \\
			\hline
			\hline
		\end{tabular}
		\caption{\label{tab:dataset}Sentences Statistics on Corpora.}
	\end{center}
	
\end{table}

Three methods reported in \cite{mikolov2013exploiting} are used as baselines: \emph{Edit Distance} (ED), \emph{Word Co-occurrence} (WC) and \emph{Translation Matrix} (shown as Mikolov in Tables \ref{tab:word-trans-wmt} and \ref{tab:word-trans-iwslt}) methods, together with state-of-the-art bilingual word embedding BilBOWA \cite{gouws2015bilbowa} and \cite{Oshikiri-acl16}'s method\footnote{The settings of their method in WMT task is different from ours, so we only compare their performance in IWSLT task.}. Their default settings are followed. We are aware there are some other related works which also focus on this task \cite{coulmance-EtAl:2015:EMNLP,xing-EtAl:2015:NAACL-HLT,Vulic-acl16}. As our best knowledge, they did not release their codes and details implements for fair comparison, until the deadline of this submission.

\subsection{Dimension Tuning}
Figures 2 and 3 show  the dimension tuning (by the average score of four sub-tasks) experiments on development data (5K) of IWSLT-2014 tasks. Because Mikolov method needs to use this 5K data for translation matrix training, their default dimension 300 is applied.  As mentioned in Section \ref{sec:nn}, the  BSC-word matrix of PCA/CA based method is an extracted graph, so the original dimension is much smaller than NN based method, where nearly all the BSCs in the whole graph are used as input of NN models.  The best performed dimension for each model will be evaluated on test data and shown in Tables \ref{tab:word-trans-wmt} and \ref{tab:word-trans-iwslt}.

\begin{figure}[htbp]
	\centering
	\includegraphics[width=3.1in]{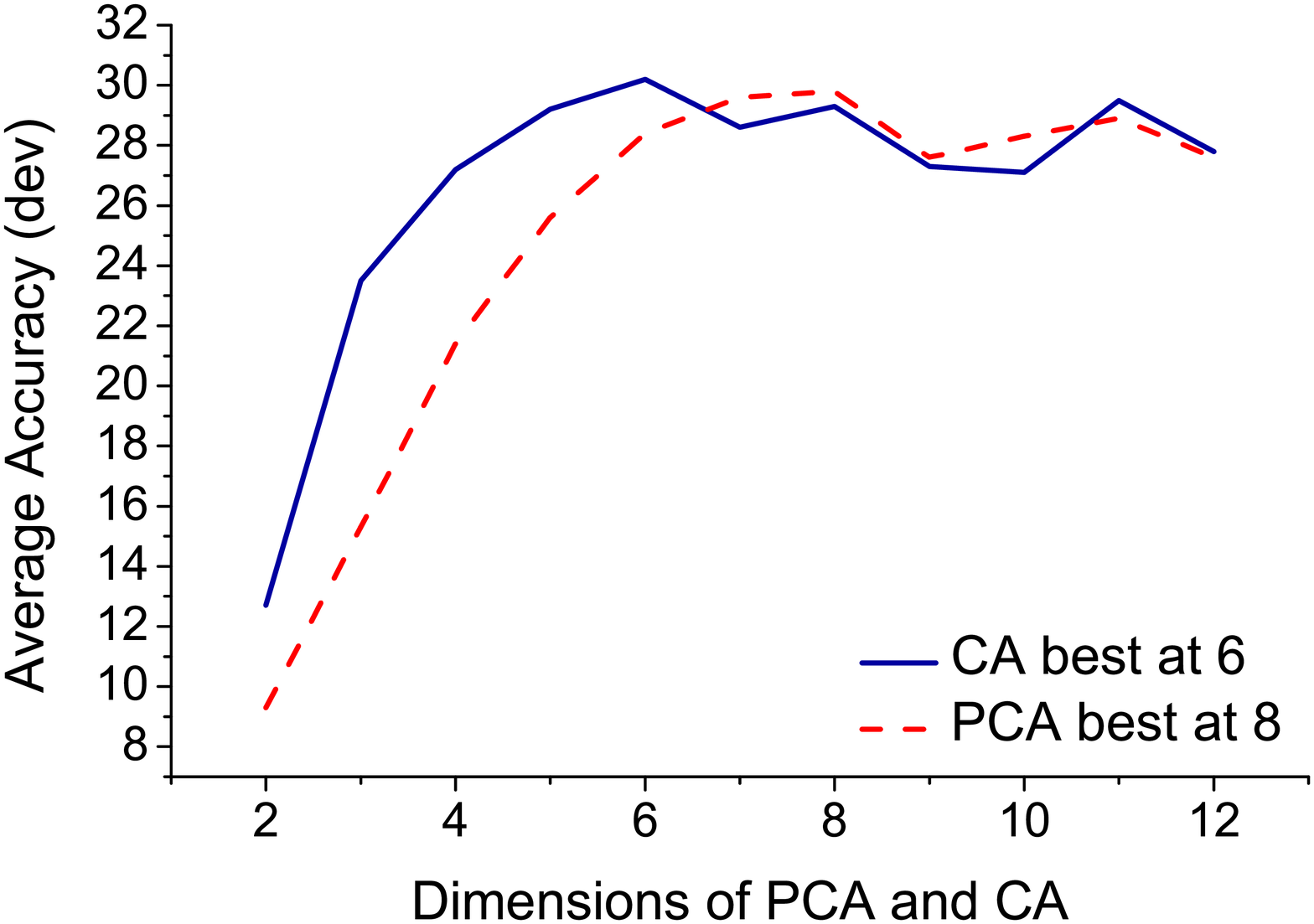}
	\includegraphics[width=3.1in]{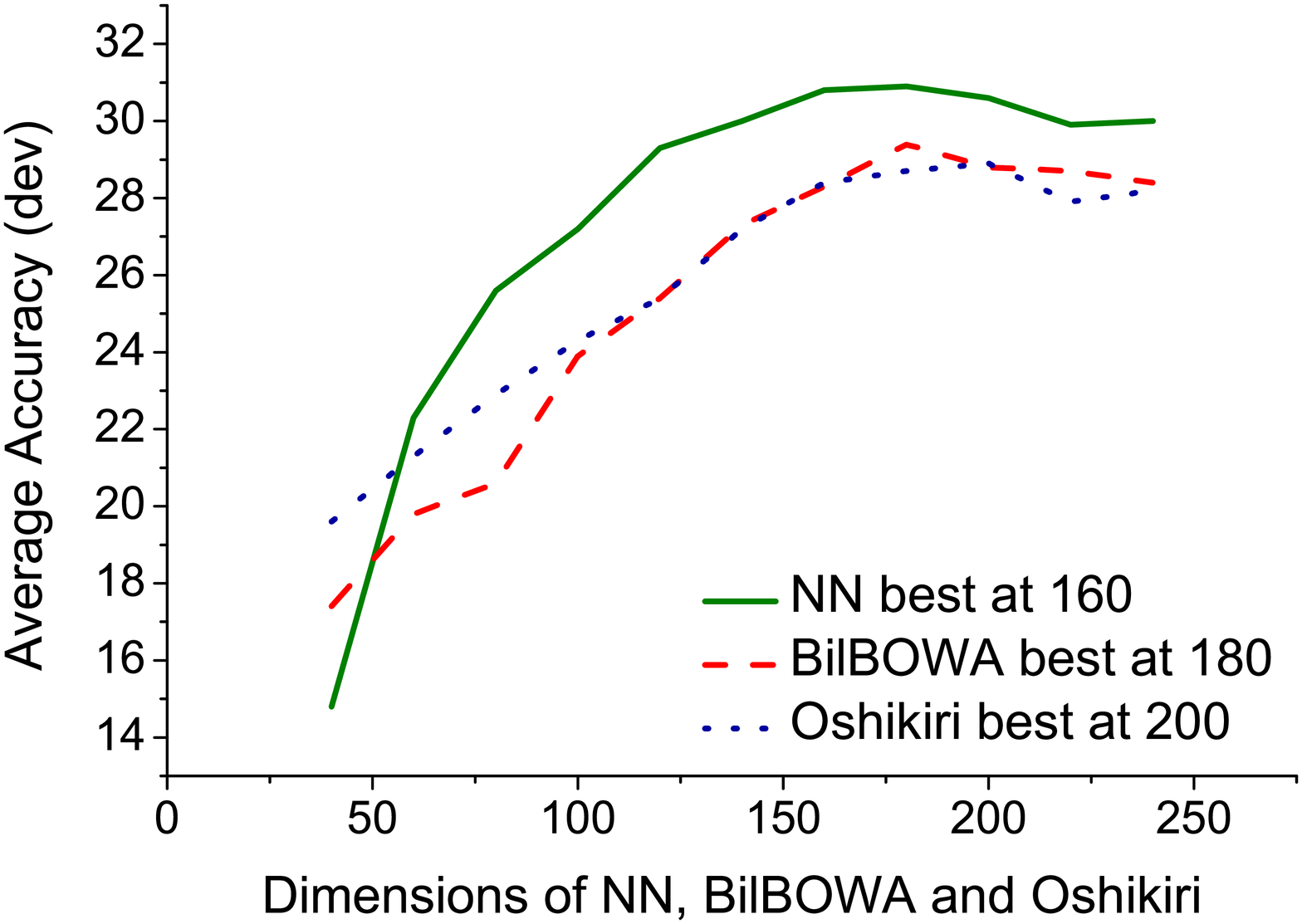}
	\caption{Dimension tuning on IWSLT task}\label{fig:tuning}
\end{figure}    

\subsection{Evaluation Results}
The evaluation results on test data are shown in Tables \ref{tab:word-trans-wmt} and \ref{tab:word-trans-iwslt}. For WMT11, the baseline results are from the reports of corresponding papers. For IWSLT14, the baseline results are from our re-implementations of corresponding methods. The results in bold indicate that they outperform corresponding best baseline results. The numbers in parentheses show how much the best results outperform the best baseline results. Similar as \cite{mikolov2013exploiting}, we also discard word pairs whose Google translation are out-of-vocabulary.

					\begin{table}[htbp]
						\begin{center}
							\begin{tabular}{l|l|l|l|l|l}
								\hline
								\hline
								WMT11  & En-Sp& Sp-En & En-Sp  & Sp-En &  Time\\
								& P@1 & P@1 & P@5 & P@5 & /hours\\
								\hline
								ED  & 13 & 18 & 24 & 27 & -\\
								WC  & 30 & 19 & 20 & 30 & -\\
								Mikolov & 33 & 35 & 51 & 52 & -\\
								BilBOWA  & 39 & \textbf{44} & 51 & 55 & -\\
								\hline
								BSC+PCA  & 34 & 40 & \textbf{55} & 53& 2.7\\
								BSC+CA  & 36 & 42 & \textbf{56(+5)} & \textbf{60(+5)}& 3.2\\
								BSC+NN  & \textbf{41(+2)} & 37 & \textbf{53} & \textbf{54}&28.7\\
								\hline
								\hline
							\end{tabular}
						\end{center}
						\caption{\label{tab:word-trans-wmt}WMT11 task.}
\end{table}

\begin{table}[htbp]
	\begin{center}
		\begin{tabular}{l|l|l|l|l|l}
			\hline
			\hline
			IWSLT14  & En-Fr & Fr-En & En-Fr  & Fr-En& Time\\
			& P@1 & P@1 & P@5 & P@5 & /hours\\
			\hline
			Mikolov & 23& 27 & 37 & 32 & 2.1\\
			BilBOWA  & 26 & 25 & 38 & 31 & 1.4\\
			Oshikiri  & 24 &  26&  32& 33& 1.6\\
			\hline
			BSC+PCA  & 18&  26&  38&  \textbf{41(+8)}& \textbf{0.9}\\
			BSC+CA  &  22&  \textbf{29(+2)}&  \textbf{41(+3)}&  \textbf{36}& 1.1\\
			BSC+NN  & \textbf{27(+1)} &  \textbf{28}&  36& \textbf{38} & 12.3\\
			\hline
			\hline
		\end{tabular}
	\end{center}
	\caption{\label{tab:word-trans-iwslt} IWSLT task.}
\end{table}

As shown in Tables \ref{tab:word-trans-wmt} and \ref{tab:word-trans-iwslt}, BSC based methods achieved the best performances in 7 out of 8 sub-tasks. The model training and calculating CPU time of BSC+PCA/CA is slightly better than the existing methods. BSC+NN is more time consuming than BSC+PCA/CA.

\section{Related Work and Discussion}
\label{sec:related}				
In this section, we will introduce the existing word embedding methods and compare them with the proposed BSE method. Word embedding for vector representation is usually built in two-steps \cite{baroni-dinu-kruszewski:2014:P14-1,schnabel-EtAl:2015:EMNLP,DBLP:journals/corr/VylomovaRCB15}. The first step is about selecting the detailed contexts related to a given word. The second step is to summary the relationship between word and its contexts into lower dimensions.

\subsection{Context Selection}	
For context selection, three categories can be identified:

1) The first category is to extract the word or word relation information from the entire text, which is usually regarded as document level processing, such as bag-of-words, Vector Space Models (VSMs), Latent Semantic Analysis \cite{Landauer97asolution}, Latent Dirichlet Allocation \cite{blei2003latent}.

2) The second category is to use sliding window, such as $n$-grams, skip-grams or other local co-occurrence  relation \cite{mikolov2013efficient,mikolov2013distributed,zou-EtAl:2013:EMNLP,levy-goldberg:2014:W14-16,pennington-socher-manning:2014:EMNLP2014}.

3) The third category, that has been seldom considered, uses much more sophisticated graph style context. Ploux and Ji \shortcite{Ploux:2003:MMS:873743.873744} describe a graph based semantic matching model using bilingual lexicons and monolingual synonyms\footnote{\url{http://dico.isc.cnrs.fr}}. They later represent words using individual monolingual co-occurrences \cite{Isc03lexicalknowledge}. Saluja et al. \shortcite{saluja-EtAl:2014:P14-1} propose a graph based method to generate translation candidates using monolingual co-occurrences. Wang et al. \shortcite{bgsm} propose bilingual graph-based semantic model which focus on Statistical Machine Translation. Oshikiri et al. propose spectral graph based cross-lingual word embeddings \cite{Oshikiri-acl16}.

Besides these methods, there are some other methods try to solve the memory and space problem in word embeddings \cite{Ling-acl16,Rothe-acl16}.

Our method extends Ploux and Ji \shortcite{Ploux:2003:MMS:873743.873744}'s monolingual method into bilingual one, which is similar with \cite{bgsm}'s work. Wang et al.  use contextual information for phrase translation and generation, and our method is for lexical translation without contextual information in comparison.

\subsection{Relationship Summarising}
For relationship summarising (dimension reduction),  Neural Networks (NNs) are very popular for word embeddings and SMT recently \cite{huang-EtAl:2012:ACL20122,mikolov2013efficient,mikolov2013distributed,zou-EtAl:2013:EMNLP,gao-EtAl:2014:P14-1,zhang-EtAl:2014:P14-11,sundermeyer-EtAl:2014:EMNLP2014,devlin-EtAl:2014:P14-1,gao-EtAl:2014:P14-1,lauly2014autoencoder,sutskever2014sequence,cho-EtAl:2014:EMNLP2014}. Besides NN methods, there are also some works which use matrix factorization \cite{pennington-socher-manning:2014:EMNLP2014,shi-EtAl:2015:ACL-IJCNLP1}, such as Correspondence Analysis (CA) \cite{Ploux:2003:MMS:873743.873744,Isc03lexicalknowledge}, Principal Component Analysis (PCA) \cite{lebret-collobert:2014:EACL} and canonical correlation analysis \cite{faruqui-dyer:2014:EACL,lu-EtAl:2015:NAACL-HLT} for word embedding.

Most of the above existing method only apply one dimension reduction method. For the proposed method, it can work together with PCA, CA and NN methods for BSC-word relationship summarising.

\subsection{Multi-sense Representations}
As we know, a word may belong to various senses (polysemous). There are several work focus on sense-specific word embeddings. Guo et al. \shortcite{guo-EtAl:2014:Coling} propose an NN based recurrent neural network based word embedding method, which makes use of previous contextual  word information. Jauhar et al. \shortcite{jauhar-dyer-hovy:2015:NAACL-HLT} use semantic vector space models for multi-sense representation learning. {\v{S}}uster et al. \shortcite{vsuster2016bilingual} learn multi-sense embeddings with discrete autoencoders by both monolingual and bilingual information.  Iacobacci et al. propose SensEmbed, which can be applied to  both word and relational similarity \cite{iacobacci-pilehvar-navigli:2015:ACL-IJCNLP,iacobacci-2}.

The key to distinguish word sense is the contextual information, which is applied to nearly all of the above methods. For the proposed method, we only consider single word translation, so we do not make use of contextual information.

\section{Conclusion}
We propose a bilingual sense unit Bilingual Sense Clique (BSC), which treats source and target words equally for embedding. Several dimension reduction methods are empirically  evaluated for summarizing the BSC-word relationship into low-dimension vectors. Empirical results on lexical translation show that the proposed methods can outperform existing bilingual embedding methods.
\bibliographystyle{acl}
\bibliography{thesis}

\end{document}